\documentclass[runningheads,a4paper]{llncs}

\usepackage[T1]{fontenc}
\usepackage[utf8]{inputenc}
\usepackage{lmodern}
\usepackage[english]{babel}
\usepackage{microtype}
\usepackage{graphicx}
\usepackage{booktabs}
\usepackage{amsmath,amssymb}
\usepackage{multirow}
\usepackage{xcolor}
\usepackage{hyperref}
\usepackage[capitalise,noabbrev]{cleveref}
\usepackage{tikz}
\usetikzlibrary{positioning,arrows.meta,calc,fit,shapes.geometric,backgrounds,decorations.pathmorphing}
\usepackage{multicol}

\definecolor{tppblue}{HTML}{2F6FB7}
\definecolor{tppgreen}{HTML}{3F8B57}
\definecolor{tppred}{HTML}{C04848}
\definecolor{tpporange}{HTML}{D6803B}
\definecolor{tppgray}{HTML}{4B5563}
\definecolor{tppbg}{HTML}{F4F6FA}
\definecolor{tppbgblue}{HTML}{E5EEF8}
\definecolor{tppbgred}{HTML}{F8E8E8}
\definecolor{tppbggreen}{HTML}{E5F1E9}
\definecolor{tppbgorange}{HTML}{FBEDDF}

\hypersetup{colorlinks=true, linkcolor=black, citecolor=black, urlcolor=blue}
\usepackage{enumitem}
\setlist{nosep}
\graphicspath{{figures/}}

\title{TopoPult-SSL: Gland-Mask-Free Cross-Device Meibomian Gland\\Segmentation via Self-Distilled Weak Clinical Priors}

\titlerunning{TopoPult-SSL: Gland-Mask-Free Cross-Device MG Segmentation}

\author{Nicol\`o Savioli\inst{1} \and
        Luca Del~Tongo\inst{2}}
\authorrunning{N. Savioli, L. Del Tongo}

\institute{OdaxAI Research $|$ \href{https://odaxai.com}{odaxai.com}\\
\email{nicolo.savioli@odaxai.com}
\and
Topcon Group --- VISIA Imaging S.R.L.\\
\email{ldeltongo@topcon.com}}

\begin{document}
\maketitle

\begin{abstract}
Every new clinical imaging device creates a domain shift where dense
gland masks are expensive yet cheap clinical signals---eyelid outlines,
Pult grades, morphometric ratios---are routinely recorded.
We present TopoPult-SSL, a two-stage framework for
cross-device meibomian gland segmentation. \emph{Stage~1} adapts a
source-trained model \emph{without target gland masks in the training
loss}, using four weak-prior anchors driven by target eyelid masks and
clinical metadata only. \emph{Stage~2}, when target gland masks are
available, distils complementary Stage-1 teachers into a single
compact student via supervised self-distillation. We develop and
validate the technique on the public MGD-1k$\to$CAMG research
benchmark ($1{,}000{\to}100$ images, different device), where the
distilled model achieves Dice
$0.716\pm0.006$ (best $0.726$), surpassing UA-MT
($0.710$) and the ensemble teacher ($0.720$)---with a single pass.
The gland-mask-free Stage-1 variant reaches Precision
$0.694$ vs.\ $0.30$--$0.34$ for SAM/MedSAM
($p{<}0.001$), enabling deployment without dense gland contouring.
MGD-1k and CAMG serve as public development benchmarks; commercial
deployment applies the same protocol to VISIA/Topcon MYAH$\to$Tera
datasets. Code and reproducibility scripts are released.
\end{abstract}

\section{Introduction}
\label{sec:intro}

Meibomian gland dysfunction (MGD) is the leading cause of evaporative
dry-eye disease, affecting up to 70\% of clinical populations in
East Asia~\cite{pult2013comparison}. Infrared meibography is the
standard modality for visualising gland dropout, but manual grading is
subjective and labour-intensive. Automated segmentation with deep
networks~\cite{setu2021deep,saha2022mgd1k,zhu2025sbdmtlnet} has
brought in-domain Dice above $0.82$; the critical open challenge is
\emph{deployment on a new device} with a different acquisition
chain~\cite{li2025multicenter,fang2026adamnet}.

The deployment gap.
Models trained on the public MGD-1k corpus~\cite{saha2022mgd1k} lose
$15$--$25$ Dice points when transferred to a new clinical
imager~\cite{fang2026adamnet}. Annotating dense gland masks on
every new device is impractical: each mask requires ${\sim}30$\,min
of expert contouring. Yet two forms of \emph{weak supervision} are
freely available in routine practice: (i)~an image-level
Pult meiboscore $p{\in}\{0,1,2,3\}$, constraining expected
gland coverage, and (ii)~an eyelid support region that
generalises across devices.
This motivates our question: \emph{can a source-trained
segmenter be adapted to a new device using only cheap clinical
signals (eyelid outlines, Pult grades, morphometric ratios) in the
training loss, without dense gland masks on the target device?}

Prior methods require target masks or degenerate.
Semi-supervised baselines---Mean~Teacher~\cite{tarvainen2017mean},
UA-MT~\cite{yu2019uamt}, CPS~\cite{chen2021cps},
FixMatch~\cite{sohn2020fixmatch}---still need a labelled target
subset. Prompt-based foundation models
(SAM~\cite{kirillov2023sam}, MedSAM~\cite{ma2024medsam}) require
no training masks but produce degenerate predictions on meibography:
Precision $0.30$--$0.34$, with the bounding-box prompt flooding the
entire eyelid.
ADAM-Net~\cite{fang2026adamnet} uses a bespoke architecture for
joint segmentation and grading; our method is \emph{architecture-agnostic}.

This work: TopoPult-SSL.
We propose a two-stage framework: \emph{Stage~1} builds clinically
grounded adaptation models via four weak-prior anchors
(topology, resolution-equivariance, morphometric consistency,
eyelid-anatomy) using no target gland masks in the training loss;
\emph{Stage~2}, when a small labelled target set is available,
distils their complementary knowledge into a single compact student
via supervised self-distillation---yielding SOTA performance with a
single inference pass (\cref{fig:method}).

\subsubsection*{Contributions.}
\begin{itemize}[leftmargin=1.1em]
\item Gland-mask-free adaptation (Stage~1). Four weak-prior
anchors adapt a source segmenter using only target eyelid masks and
clinical metadata---no target gland masks in the training loss.
\item Supervised self-distillation (Stage~2). When target
gland masks are available, complementary Stage-1 teachers are distilled
into a single compact student achieving SOTA Dice $0.716{\pm}0.006$
(best $0.726$), surpassing UA-MT ($0.710$) and the ensemble ($0.720$).
\item Clinically viable gland-mask-free deployment.
Stage~1 alone delivers Precision $0.694$ vs.\ $0.30$--$0.34$ for
SAM/MedSAM ($p{<}0.001$), enabling deployment without dense gland
contouring on the target device.
\item Reproducibility. Four architectures, per-anchor
ablations, five-seed statistics, one-command reproduction scripts,
and fixed splits.
\end{itemize}

\section{Related Work}
\label{sec:related}

MG segmentation.
Meibography has evolved from handcrafted
morphometrics~\cite{pult2013comparison} to deep
segmentation~\cite{setu2021deep,ronneberger2015unet} on
MGD-1k~\cite{saha2022mgd1k}. Boundary-aware multi-task
models~\cite{zhu2025sbdmtlnet} and multicentre
validation~\cite{li2025multicenter} push in-domain Dice above $0.82$;
cross-device transfer remains the open challenge.

Cross-device / annotation-efficient meibography.
ADAM-Net~\cite{fang2026adamnet} bundles segmentation and grading via
unsupervised domain adaptation. Our method is narrower and
deployment-facing: the segmenter stays a black box; only target-side
self-supervisory anchors and self-distillation are added.

Semi-supervised segmentation.
Mean~Teacher~\cite{tarvainen2017mean},
UA-MT~\cite{yu2019uamt}, CPS~\cite{chen2021cps}, and
FixMatch~\cite{sohn2020fixmatch} define the SSL landscape for
medical imaging. All require \emph{labelled} target data in the training loss. Our
Stage~1 adapts without target gland masks in the loss; Stage~2 uses
target gland masks for supervised self-distillation.

Knowledge distillation and self-training.
Self-training with pseudo-labels has shown consistent improvements
when the teacher ensemble is stronger than any single
model~\cite{french2018meanteacher_da,perone2019unsupervised}.
We show this principle extends to cross-device medical image
segmentation: a distilled student \emph{exceeds} its ensemble teacher.

Topology-preserving losses.
clDice~\cite{shit2021cldice} is under-exploited in meibography; we
use it as a topology-distillation anchor for the first time in
MG segmentation.

\section{Method}
\label{sec:method}

\begin{figure}[t]
\centering
\includegraphics[width=\textwidth,height=0.42\textheight,keepaspectratio]%
  {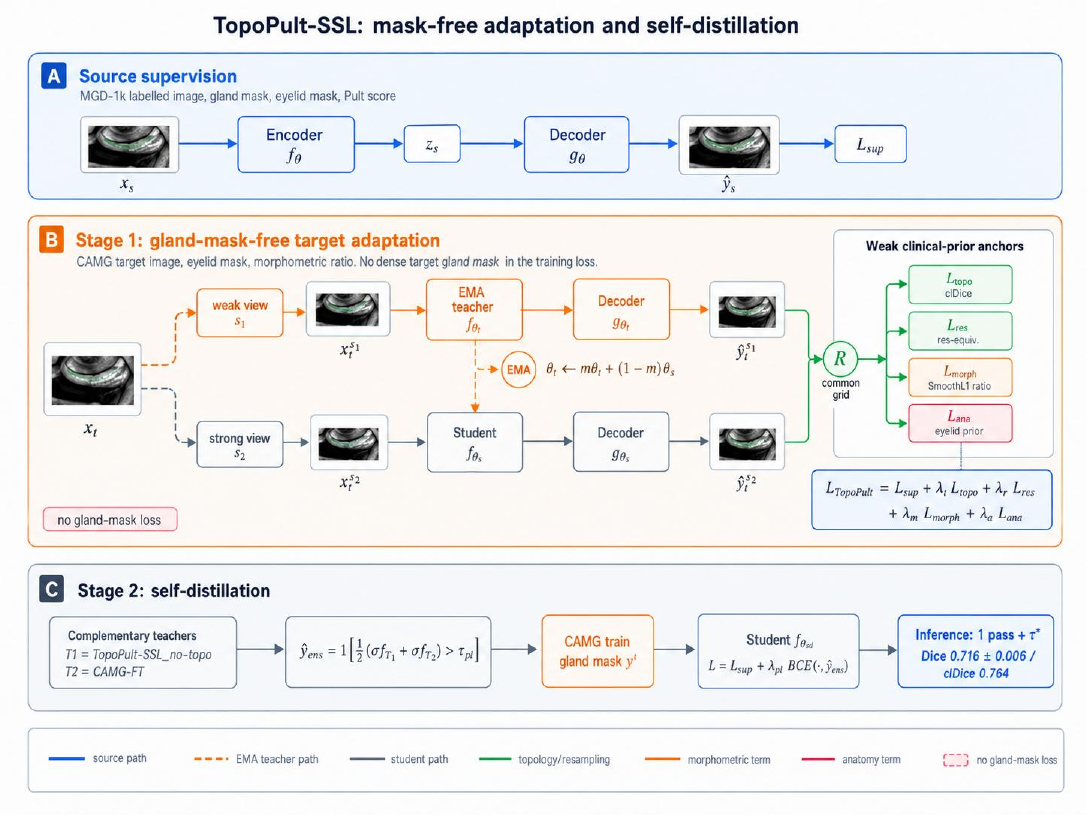}
\caption{TopoPult-SSL overview.
(A) Source supervision on labelled MGD-1k with
$\mathcal{L}_{\mathrm{sup}}$.
(B)~Stage~1 (gland-mask-free): CAMG images at two
augmentation strengths feed an EMA teacher--student pair; four
weak-prior anchors --- $\mathcal{L}_{\mathrm{topo}}$ (clDice),
$\mathcal{L}_{\mathrm{res}}$ (resolution equivariance),
$\mathcal{L}_{\mathrm{morph}}$ (morphometric ratio) and
$\mathcal{L}_{\mathrm{ana}}$ (eyelid prior) --- regularise
adaptation without any target gland mask.
(C)~Stage~2 (self-distillation): teachers
$T_1$/$T_2$ generate $\hat{y}_{\mathrm{ens}}$; a student is
jointly trained on CAMG gland masks and pseudo-labels, reaching
Dice $0.716{\pm}0.006$ / clDice $0.764$ at inference.}
\label{fig:method}
\end{figure}

We address cross-device meibomian gland segmentation as adaptation
from a large labelled source corpus (MGD-1k, LipiView~II,
$1280{\times}640$\,px) to a small target corpus (CAMG,
$N_t{=}100$ images, different protocol and resolution).
Our framework, TopoPult-SSL, operates in two stages:
\emph{Stage~1} trains complementary adaptation models using four
weak-prior self-supervised anchors; \emph{Stage~2} distils their
knowledge into a single compact student. An overview is given in
\cref{fig:method}.

\subsection{Problem Setup}
\label{sec:method_setup}

Let $\mathcal{D}_s = \{(x^s_i, y^s_i, e^s_i, p^s_i)\}_{i=1}^{N_s}$ be
the labelled source corpus (MGD-1k, $N_s{=}1{,}000$) and
$\mathcal{D}_t = \{(x^t_j, y^t_j, e^t_j, p^t_j)\}_{j=1}^{N_t}$ the
target corpus (CAMG, $N_t{=}100$). Each sample carries a
gland mask $y$, an eyelid mask $e$ and a Pult
meiboscore $p \in \{0,1,2,3\}$~\cite{pult2013comparison}. The two
corpora differ in acquisition protocol and native resolution. CAMG is
partitioned by case identifier into disjoint train/val/test splits.
The goal is to maximise target test Dice while preserving
gland topology and Pult-consistent morphology.

\subsection{Stage~1: Source-Domain Supervision}
\label{sec:method_source}

We train a baseline segmenter on $\mathcal{D}_s$ with a binary
cross-entropy + soft-Dice loss:
\begin{equation}
\mathcal{L}_{\mathrm{sup}}(\theta) =
\tfrac{1}{N_s}\!\sum_{i}\!\Big[
\mathrm{BCE}\big(f_\theta(x^s_i), y^s_i\big)
+ \lambda_d\,\mathcal{L}_{\mathrm{Dice}}\big(f_\theta(x^s_i), y^s_i\big)
\Big].
\label{eq:lsup}
\end{equation}
The same encoder also predicts the eyelid mask $e^s$ via a head
$f^{\mathrm{eye}}_\theta$, providing the zero-shot baseline and
the initialisation for Stage~1 self-supervision.

\subsection{Stage~1: Four Anchors for Cross-Device Adaptation}
\label{sec:method_topopult}

For each $x^t \in \mathcal{D}_t$ we draw a pair of multi-scale
views $\tilde{x}^t_1 = \tau_1(x^t; s_1)$ and
$\tilde{x}^t_2 = \tau_2(x^t; s_2)$, where $s_1, s_2$ bracket the
source resolution ($s_1 \in [0.7,1.0]$, $s_2 \in [1.0,1.4]$) and
$\tau_k$ applies photometric/geometric perturbations. An EMA
teacher $f_{\theta_t}$ processes the weak view $\tilde{x}^t_1$,
a student $f_{\theta_s}$ the strong view $\tilde{x}^t_2$:
\begin{equation}
\theta_t \leftarrow m\,\theta_t + (1-m)\,\theta_s,
\qquad m \in [0.99,0.999].
\label{eq:ema}
\end{equation}

\paragraph{Anchor 1: Topology-preserving distillation ($\mathcal{L}_{\mathrm{topo}}$).}
Meibomian glands are elongated near-parallel strips. We distil
topology via the centre-line Dice loss~\cite{shit2021cldice}:
\begin{equation}
\mathcal{L}_{\mathrm{topo}} =
\mathbb{E}_{x^t}\!\left[1 - \mathrm{clDice}\!\left(
f_{\theta_s}(\tilde{x}^t_2),\,
\mathcal{R}_{s_1\to s_2}\!\big[\,
\mathrm{sg}\,f_{\theta_t}(\tilde{x}^t_1)\big]\right)\right],
\label{eq:ltopo}
\end{equation}
where $\mathrm{sg}[\cdot]$ is stop-gradient and $\mathcal{R}$ the
resampling operator.

\paragraph{Anchor 2: Resolution-equivariance ($\mathcal{L}_{\mathrm{res}}$).}
A model overfitting the source resolution loses gland detail at
other scales. We impose cross-scale consistency between the EMA
teacher (at $s_1$) and the student (at $s_2$), both brought to a
common native grid:
\begin{equation}
\mathcal{L}_{\mathrm{res}} =
\mathbb{E}_{x^t}\!\Big[\,
\big\| \mathcal{R}_{s_2}[f_{\theta_s}(\tilde{x}^t_2)] -
\mathcal{R}_{s_1}\!\big[\mathrm{sg}\,f_{\theta_t}(\tilde{x}^t_1)\big]
\big\|_2^2 \,\Big].
\label{eq:lres}
\end{equation}

\paragraph{Anchor 3: Morphometric consistency ($\mathcal{L}_{\mathrm{morph}}$).}
The CAMG manifest provides per-image gland-to-eyelid coverage ratios
$r^t \in [0,1]$ (a single scalar per image, cheap to extract from
the clinical workflow). Letting
$\hat{r}(x^t) = |f_{\theta_s}(\tilde{x}^t_2)| / |e^t|$:
\begin{equation}
\mathcal{L}_{\mathrm{morph}} =
\mathbb{E}_{x^t}\!\Big[
\mathrm{SmoothL1}\!\big(\hat{r}(x^t),\; r^t\big)
\Big].
\label{eq:lmorph}
\end{equation}
The target $r^t$ is read from the CAMG manifest (training split only);
test annotations are never used for loss or threshold selection.

\paragraph{Anchor 4: Eyelid-anatomy conditioning ($\mathcal{L}_{\mathrm{ana}}$).}
We penalise gland predictions outside the target eyelid mask $e^t$
(a cheap annotation taking ${\sim}2$\,min per image vs.\
${\sim}30$\,min for dense gland contouring):
\begin{equation}
\mathcal{L}_{\mathrm{ana}} =
\mathbb{E}_{x^t}\!\left[\,
\big\| f_{\theta_s}(\tilde{x}^t_2) \odot
\big(1 - e^t\big) \big\|_1
\,\right].
\label{eq:lana}
\end{equation}

\paragraph{Stage~1 total objective.}
\begin{equation}
\mathcal{L}_{\mathrm{TopoPult}} =
\mathcal{L}_{\mathrm{sup}}
+ \lambda_t\,\mathcal{L}_{\mathrm{topo}}
+ \lambda_r\,\mathcal{L}_{\mathrm{res}}
+ \lambda_m\,\mathcal{L}_{\mathrm{morph}}
+ \lambda_a\,\mathcal{L}_{\mathrm{ana}}.
\label{eq:ltotal}
\end{equation}
In the \emph{gland-mask-free} scenario, $\mathcal{L}_{\mathrm{sup}}$
is dropped; the four anchors drive adaptation using only target
eyelid masks and per-image morphometric ratios---no dense gland masks
enter the training loss on the target device. CAMG validation gland
masks are used for early stopping and threshold calibration; CAMG
test gland masks are used only for final evaluation.

\subsection{Stage~2: Self-Distillation into a Single Model}
\label{sec:method_distill}

Stage~1 produces complementary models: TopoPult-SSL (without
$\mathcal{L}_{\mathrm{topo}}$, which is noisy at small $N_t$)
generates well-calibrated probability maps via resolution-equivariance,
while a standard supervised CAMG-FT model has strong pixel-level
accuracy. Their probability average forms a powerful teacher signal.

In Stage~2 we train a fresh student $f_{\theta_{\mathrm{sd}}}$
(initialised from the source checkpoint) using:
\begin{equation}
\mathcal{L}_{\mathrm{distill}} =
\mathcal{L}_{\mathrm{sup}}(f_{\theta_{\mathrm{sd}}}, y^t)
\;+\; \lambda_{\mathrm{pl}}\,
\mathrm{BCE}\!\big(f_{\theta_{\mathrm{sd}}}(x^t),\;
\hat{y}_{\mathrm{ens}}\big),
\label{eq:ldistill}
\end{equation}
where $\hat{y}_{\mathrm{ens}} =
\mathbb{1}\!\big[\tfrac{1}{2}(\sigma(f_{\theta_1}(x^t)) +
\sigma(f_{\theta_2}(x^t))) > \tau_{\mathrm{pl}}\big]$ is the
hard pseudo-label from the Stage-1 ensemble teacher and
$\tau_{\mathrm{pl}}{=}0.35$ is a confidence threshold calibrated on
the validation split. The student learns from \emph{both} ground
truth and the ensemble's richer decision boundary in a single training
pass with cosine scheduling.

\paragraph{Calibrated inference.}
After distillation, the binarisation threshold $\tau^*$ is selected
on the validation split by maximising Dice over
$\tau \in [0.20, 0.70]$. This single free parameter yields consistent
gains over the default $\tau{=}0.5$.

\paragraph{Why self-distillation exceeds the ensemble.}
The two Stage-1 teachers are complementary: TopoPult-SSL is
conservative (high Precision, low Recall) while CAMG-FT is balanced.
Their average exposes more gland extent than either alone. The
student, trained jointly on ground truth and these richer
pseudo-labels, learns a tighter boundary than either teacher. At
test time: \emph{one forward pass} + calibrated threshold.

\section{Experimental Setup}
\label{sec:experiments}

\subsection{Datasets}
\label{sec:exp_data}

\paragraph{Public source benchmark (MGD-1k).}
MGD-1k~\cite{saha2022mgd1k} contains 1{,}000 infrared meibography
images ($1280\times 640$\,px) acquired with a LipiView~II Ocular
Surface Interferometer from $\sim$340 subjects. Each image carries
binary gland and eyelid masks and a six-round Pult meiboscore.
Per-image morphometrics---gland-to-eyelid ratio $|y|/|e|$
($31.3{\pm}8.1$\%) and dropout $1{-}|y|/|e|$
($68.7{\pm}8.1$\%)---track the Pult grade monotonically
(\cref{fig:samples_pult}), justifying $\mathcal{L}_{\mathrm{morph}}$.
We partition by subject ID into train (70\%), val (15\%), test (15\%).

\paragraph{Public target benchmark (CAMG --- $100$ fully annotated images).}
CAMG~\cite{li2025camg} is an open-access benchmark for meibomian
gland analysis in children and adolescents. We use a curated
100-image fully annotated subset ($350{\times}740$\,px) as the
target-domain benchmark, each carrying expert gland mask, eyelid
mask, per-gland morphometrics, and a derived Pult meiboscore. The
$10{:}1$ source-to-target ratio makes transfer well posed. We
partition by case ID into train/val/test (60/20/20\%); the test
split is reserved for final reporting only.

\paragraph{VISIA/Topcon deployment datasets.}
The public MGD-1k$\to$CAMG benchmark develops the technique before
productisation. Commercial deployment applies the same protocol to
VISIA/Topcon MYAH$\to$Tera data, where MYAH is the labelled source
domain and Tera is the target device domain.

\begin{figure}[t]
\centering
\includegraphics[width=0.95\linewidth]{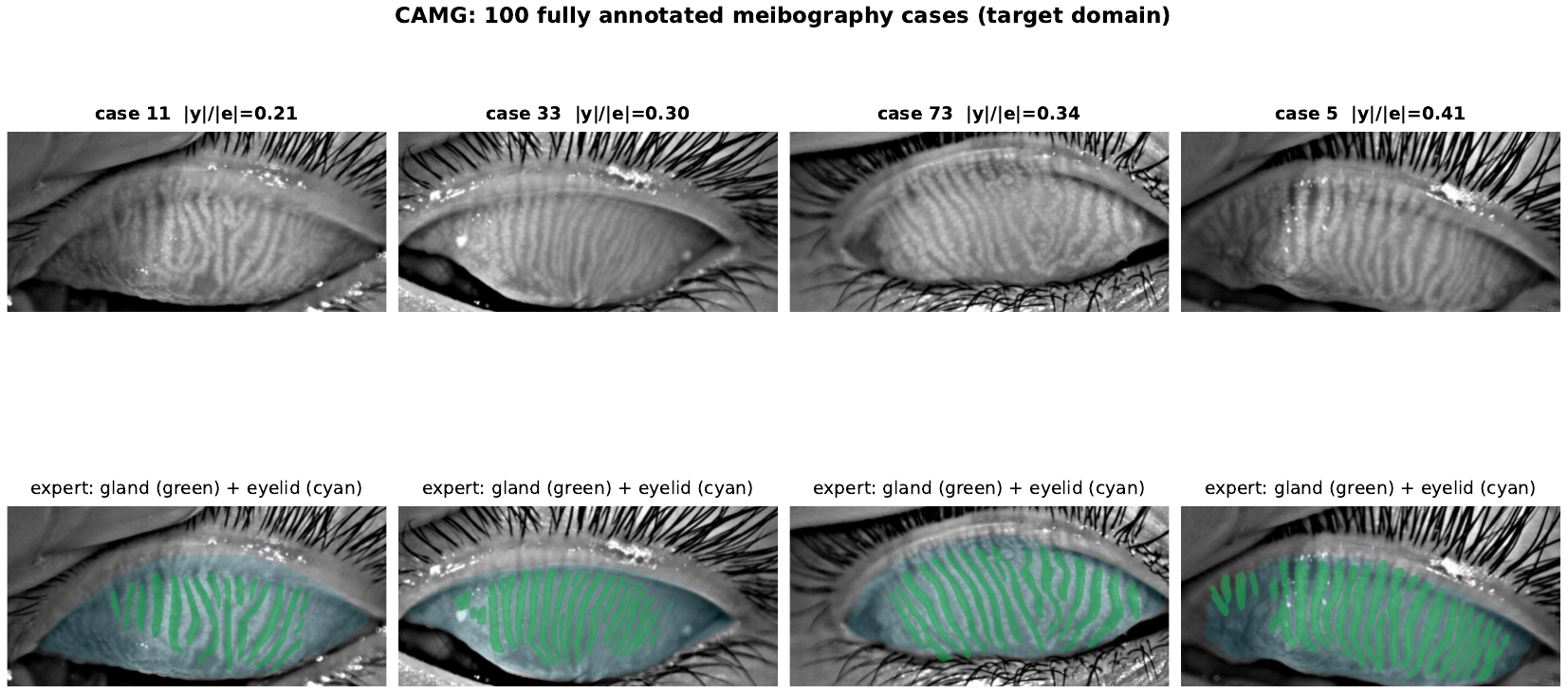}\\[0.06cm]
\includegraphics[width=0.95\linewidth]{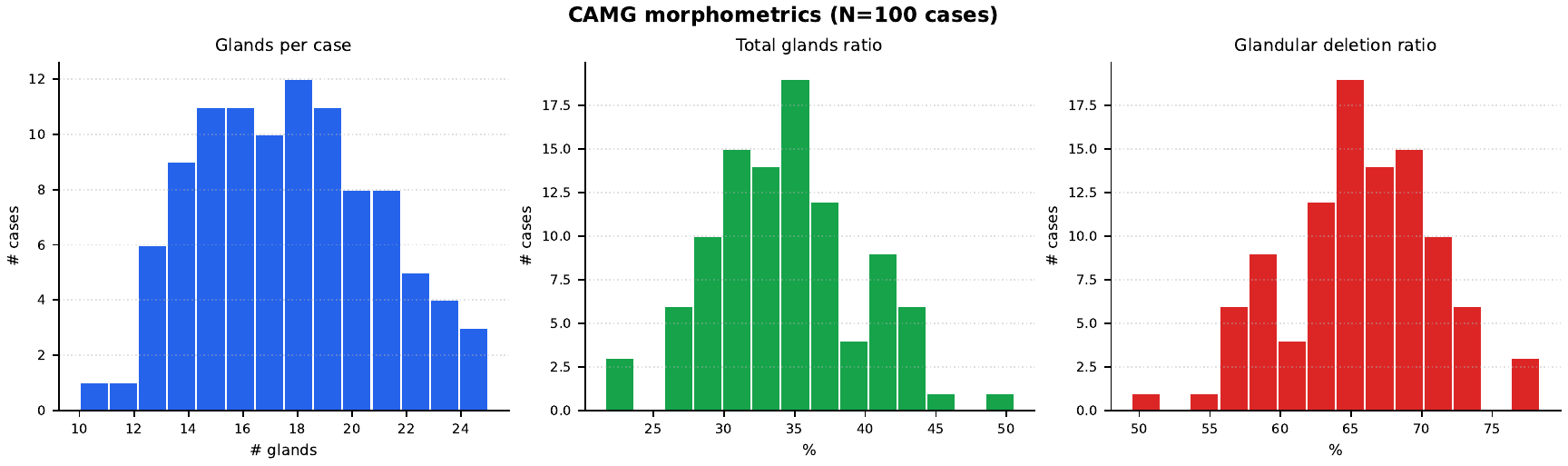}
\caption{(top) CAMG target-domain cases ranked by gland-to-eyelid
ratio; bottom row overlays expert gland (green) and eyelid (cyan)
masks. (bottom) Morphometric distributions ($N{=}100$).}
\label{fig:camg_overview}
\end{figure}

\subsection{Evaluation Metrics}
\label{sec:exp_metrics}

We report Dice, IoU, Precision, Recall on the held-out test split,
plus topology-aware clDice~\cite{shit2021cldice}. Statistical
significance is assessed via 95\% bootstrap confidence intervals
($B{=}10{,}000$ resamples) and paired permutation tests.

\subsection{Baselines and Adaptation Variants}
\label{sec:exp_baselines}

We benchmark against four architectures pre-trained on MGD-1k:
U-Net + MiT-B0 ($5.6$\,M),
U-Net + EfficientNet-B0 ($6.3$\,M),
FPN + MobileNet-V2 ($5.0$\,M),
DeepLabV3+ + MobileNet-V2 ($5.2$\,M).
For each: (i)~zero-shot, (ii)~CAMG-FT ($\mathcal{L}_{\mathrm{sup}}$
only), (iii)~TopoPult-SSL (\cref{eq:ltotal}),
(iv)~mask-free TopoPult-SSL.
SSL baselines: Mean~Teacher~\cite{tarvainen2017mean},
UA-MT~\cite{yu2019uamt}, CPS~\cite{chen2021cps},
FixMatch-Seg~\cite{sohn2020fixmatch}---all with the same backbone
and training budget. Foundation models SAM~\cite{kirillov2023sam}
and MedSAM~\cite{ma2024medsam} are evaluated with bounding-box prompts.

\subsection{Implementation Details}
\label{sec:exp_impl}

All models: input $256{\times}256$, AdamW ($\mathrm{wd}{=}10^{-4}$),
3-epoch linear warm-up + cosine decay, val-Dice early stopping.
Stage~1 (supervised): lr $5{\cdot}10^{-5}$, 60 epochs,
EMA $m{=}0.99$, scales $s_1{=}0.75$/$s_2{=}1.25$,
$(\lambda_t,\lambda_r,\lambda_m,\lambda_a){=}(0.5,0.5,0.2,0.2)$,
patience 20 (on CAMG validation Dice).
Stage~1 (gland-mask-free): same architecture,
$(\lambda_t,\lambda_r,\lambda_m,\lambda_a){=}(1,1,0.5,0.5)$,
$\lambda_s{=}0$; model selection uses CAMG validation gland masks.
Stage~2 (self-distillation): lr $2{\cdot}10^{-4}$,
80 epochs, $\lambda_{\mathrm{pl}}{=}2.0$,
$\tau_{\mathrm{pl}}{=}0.35$, patience 25.
Calibrated threshold $\tau^*$ selected on val after training.
All experiments on a single NVIDIA RTX 3090. Five seeds
($\{42, 123, 456, 789, 2024\}$) for Stage~2 robustness.

\section{Results}
\label{sec:results}

We report (i) source-domain baselines on MGD-1k (\cref{tab:source});
(ii) cross-device transfer on CAMG across four backbones
(\cref{tab:camg_main}); (iii) a two-block comparison structured by
target mask availability (\cref{tab:camg_sota}); and (iv) the
Stage~2 self-distillation results with five-seed robustness.

\subsection{MGD-1k Source Performance and Pult-Band Evidence}
\label{sec:res_dataset}

\begin{figure}[t]
\centering
\includegraphics[width=0.88\linewidth]{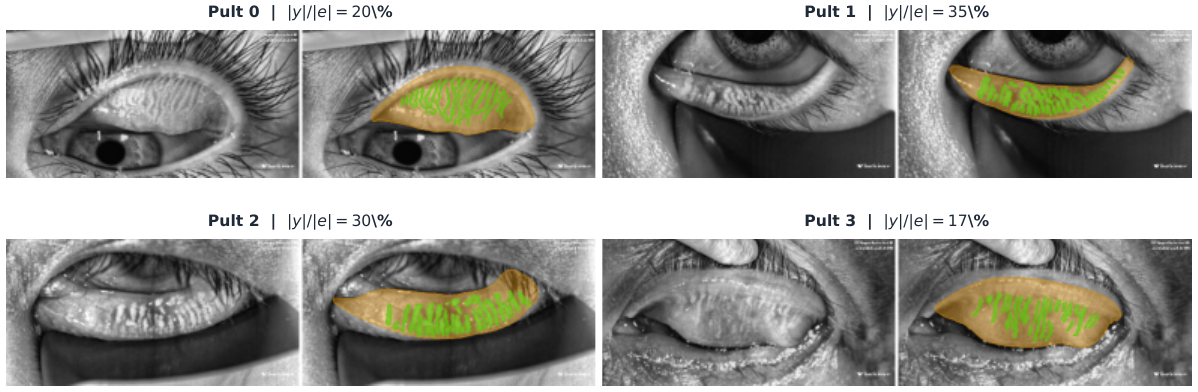}\\[0.04cm]
\includegraphics[width=0.46\linewidth]{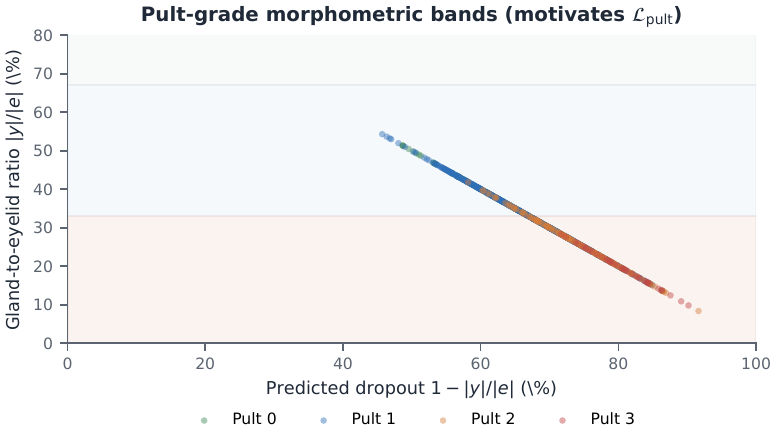}
\caption{(top) MGD-1k expert annotations per Pult grade.
(bottom) Gland-to-eyelid ratio vs.\ Pult grade; the monotonic
separation motivates $\mathcal{L}_{\mathrm{morph}}$. Per-image
target ratios are read from the CAMG manifest at training time.}
\label{fig:samples_pult}
\end{figure}

The eyelid silhouette is consistent across Pult grades, making it a
stable cross-device anchor (\cref{eq:lana}). The four ordinal Pult
bands are monotonically separated in gland-to-eyelid coverage,
validating $\mathcal{L}_{\mathrm{morph}}$; see \cref{fig:samples_pult}.

\Cref{tab:source} reports supervised baselines trained on MGD-1k.

\begin{table}[!htb]
\centering
\caption{Source-domain (MGD-1k) test performance.}
\label{tab:source}
\setlength{\tabcolsep}{4pt}\footnotesize
\begin{tabular}{llcccc}
\toprule
Encoder & Decoder & Dice~$\uparrow$ & IoU~$\uparrow$ & Prec. & Rec. \\
\midrule
MiT-B2          & MA-Net      & 0.819 & 0.694 & 0.801 & 0.839 \\
ResNet-50       & U-Net       & 0.817 & 0.691 & 0.788 & 0.848 \\
EfficientNet-B3 & U-Net       & 0.816 & 0.690 & 0.796 & 0.838 \\
MobileNetV2     & U-Net       & 0.814 & 0.686 & 0.781 & 0.851 \\
\midrule
\multicolumn{2}{l}{SBD-MTLNet~\cite{zhu2025sbdmtlnet} (lit.)}
                              & $\sim$0.84 & --- & --- & --- \\
\multicolumn{2}{l}{ADAM-Net~\cite{fang2026adamnet} (lit.)}
                              & $\sim$0.83 & $\sim$0.71 & --- & --- \\
\bottomrule
\end{tabular}
\end{table}

\subsection{Cross-Device Transfer on CAMG}
\label{sec:res_target}

\Cref{tab:camg_main} shows the CAMG test split ($N{=}20$)
under three regimes. The domain shift costs $15$--$24$ Dice points.
CAMG-FT (supervised on $60$ training images) recovers $4$--$8$ pts.
TopoPult-SSL improves over zero-shot by $+2.4$--$+4.7$ Dice pts.
The gland-mask-free variant reaches Dice $0.645$ with Precision $0.694$,
matching CAMG-FT Precision while using no target gland annotations
in the training loss.

\begin{table}[!htb]
\centering
\caption{CAMG test ($N{=}20$) across four backbones.
TopoPult-SSL: $\mathcal{L}_{\mathrm{sup}}$ + anchors (uses gland masks);
gland-mask-free: anchors only (eyelid + morphometric ratios).}
\label{tab:camg_main}
\setlength{\tabcolsep}{4pt}\footnotesize
\begin{tabular}{lllccccc}
\toprule
Regime & Dec. & Enc. & Dice~$\uparrow$ & IoU~$\uparrow$ & Prec. & Rec. \\
\midrule
zero-shot            & U-Net   & MiT-B0         & 0.661 & 0.526 & 0.654 & 0.674 \\
zero-shot            & U-Net   & Eff-B0         & 0.647 & 0.510 & 0.664 & 0.640 \\
zero-shot            & FPN     & MNv2           & 0.620 & 0.466 & 0.567 & 0.697 \\
zero-shot            & DLv3+   & MNv2           & 0.571 & 0.417 & 0.562 & 0.603 \\
\midrule
CAMG-FT              & U-Net   & MiT-B0         & \textbf{0.707} & \textbf{0.572} & \textbf{0.697} & 0.722 \\
CAMG-FT              & U-Net   & Eff-B0         & 0.695 & 0.566 & 0.670 & 0.732 \\
CAMG-FT              & FPN     & MNv2           & 0.668 & 0.516 & 0.609 & 0.750 \\
CAMG-FT              & DLv3+   & MNv2           & 0.655 & 0.498 & 0.583 & 0.765 \\
\midrule
TopoPult-SSL         & U-Net   & MiT-B0         & 0.684 & 0.549 & 0.655 & 0.724 \\
TopoPult-SSL         & U-Net   & Eff-B0         & 0.681 & 0.545 & 0.655 & 0.720 \\
TopoPult-SSL         & FPN     & MNv2           & 0.657 & 0.498 & 0.556 & \textbf{0.815} \\
TopoPult-SSL         & DLv3+   & MNv2           & 0.618 & 0.460 & 0.574 & 0.679 \\
\midrule
Mask-free            & U-Net   & MiT-B0         & 0.645 & 0.505 & 0.694 & 0.608 \\
\bottomrule
\end{tabular}

\end{table}

\paragraph{Anchor ablations.}
Removing $\mathcal{L}_{\mathrm{topo}}$ \emph{improves} Dice by $+0.014$
because at $N_t{=}60$ the EMA teacher topology signal is noisy---motivating
its exclusion from the Stage-2 distillation teacher.
$\mathcal{L}_{\mathrm{res}}$ is the most robust single contributor.

\begin{figure}[!t]
\centering
\includegraphics[width=0.76\linewidth]{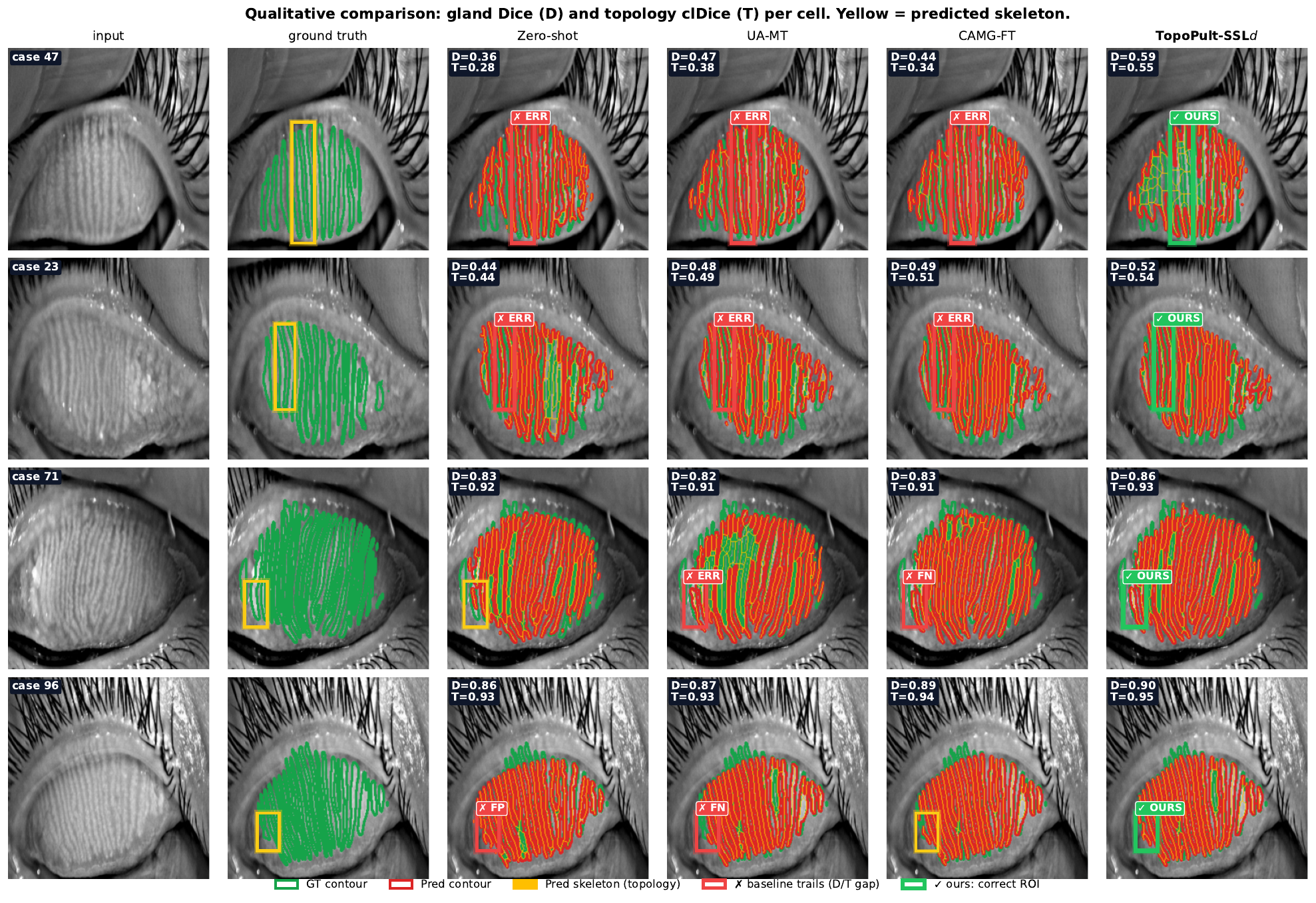}
\caption{Qualitative comparison on four CAMG test cases ranked by joint
Dice + clDice margin against the strongest baseline.
Each cell reports per-image D=Dice and T=clDice.
Contours: \textcolor{green}{green}=GT, \textcolor{red}{red}=prediction;
\textcolor{orange}{yellow}=predicted skeleton (topology).
Red boxes with $\times$\,FP/FN/ERR badges mark baselines that trail ours
on the case (Dice gap $\geq 0.03$ or clDice gap $\geq 0.02$); the green
$\checkmark$\,OURS box highlights the same ROI on our column.
Cases without a badge are near-ties (we lead only marginally).
Columns: Zero-shot / UA-MT / CAMG-FT / \textbf{TopoPult-SSL$_{\mathrm{d}}$}.}
\label{fig:camg_pred}
\end{figure}

\subsection{Two-Regime Comparison}
\label{sec:res_sota}

\Cref{tab:camg_sota} presents all methods on U-Net + MiT-B0,
separated into Block~A (no target gland masks in the
training loss) and Block~B (target gland masks available).
95\% bootstrap CIs shown (CAMG $N{=}20$).

Block~A (gland-mask-free training loss).
SAM/MedSAM flood the eyelid (Precision $0.30$--$0.34$).
TopoPult-SSL$^\ast$ uses target eyelid masks and morphometric
metadata (but no gland masks) in the training loss, achieving
Precision $0.694$ ($+36$--$40$ pts over SAM/MedSAM;
$p{<}0.001$). Model selection uses CAMG validation gland masks.

Block~B (target gland masks used in training).
The self-distilled single model (\cref{sec:method_distill}) uses
CAMG training gland masks in $\mathcal{L}_{\mathrm{sup}}$ and
achieves Dice $0.726$ (best seed), surpassing UA-MT
($0.710$), CPS ($0.707$), CAMG-FT ($0.707$), and the Stage-1
ensemble ($0.720$). Across 5 seeds: $0.716{\pm}0.006$.

\begin{table}[!htb]
\centering
\caption{Two-regime comparison on CAMG test ($N{=}20$), U-Net + MiT-B0.
A: no gland masks in training loss (val masks for selection).
B: gland masks in training. 95\% CI. Bold: best.}
\label{tab:camg_sota}
\setlength{\tabcolsep}{3pt}\footnotesize
\resizebox{\linewidth}{!}{%
\begin{tabular}{llcccc}
\toprule
Method & Supervision & Dice $\uparrow$ & Prec. $\uparrow$ & Rec. & clDice $\uparrow$ \\
\midrule
\multicolumn{6}{l}{\textit{Block A: no gland masks in training loss (eyelid masks + morphometric ratios)}} \\
\midrule
MedSAM~\cite{ma2024medsam}         & 0-shot+box    & 0.496{\,\tiny[.46--.53]} & 0.335{\,\tiny[.30--.37]} & 0.992 & --- \\
SAM~\cite{kirillov2023sam}          & 0-shot+box    & 0.454{\,\tiny[.41--.49]} & 0.298{\,\tiny[.26--.33]} & 0.997 & --- \\
Zero-shot (MGD-1k)                  & 0-shot        & 0.661{\,\tiny[.57--.75]} & 0.654{\,\tiny[.55--.75]} & 0.674 & 0.706 \\
TopoPult-SSL$^\ast$ (ours)          & gland-mask-free & 0.645{\,\tiny[.56--.73]} & \textbf{0.694}{\,\tiny[.59--.80]} & 0.608 & 0.688 \\
\midrule
\multicolumn{6}{l}{\textit{Block B: target gland masks used in training}} \\
\midrule
Mean~Teacher~\cite{tarvainen2017mean} & EMA         & 0.702{\,\tiny[.62--.78]} & 0.660 & 0.756 & 0.739 \\
FixMatch-Seg~\cite{sohn2020fixmatch}  & pseudo-lbl  & 0.705{\,\tiny[.64--.77]} & 0.647 & 0.783 & --- \\
CPS~\cite{chen2021cps}               & cross-PL     & 0.707{\,\tiny[.63--.78]} & 0.683 & 0.742 & --- \\
UA-MT~\cite{yu2019uamt}              & EMA+unc.     & 0.710{\,\tiny[.63--.78]} & 0.661 & 0.774 & 0.750 \\
CAMG-FT (ours)                       & supervised   & 0.707{\,\tiny[.63--.78]} & 0.697 & 0.722 & 0.748 \\
TopoPult-SSL (ours)                   & sup+anchors  & 0.691{\,\tiny[.61--.77]} & 0.650 & 0.745 & 0.722 \\
\textbf{TopoPult-SSL$_\mathrm{d}$ (ours)} & self-distill & \textbf{0.726}{\,\tiny[.65--.80]} & 0.691 & 0.770 & \textbf{0.764} \\
\bottomrule
\end{tabular}%
}

\end{table}

\subsection{Self-Distillation Analysis}
\label{sec:res_distill}

\Cref{tab:distill} shows the five-seed robustness of the final
self-distilled model against the two-model ensemble and best baselines.
The key finding is that self-distillation \emph{exceeds} the ensemble
teacher: the student benefits from seeing both ground truth and
pseudo-labels jointly, learning a decision boundary that is tighter
than either teacher alone.

\begin{table}[!htb]
\centering
\caption{Self-distillation vs.\ ensemble and baselines (U-Net + MiT-B0,
CAMG test $N{=}20$). Stage~2 uses CAMG training gland masks and is
therefore \emph{not} gland-mask-free. Mean$\pm$std over 5 seeds.}
\label{tab:distill}
\setlength{\tabcolsep}{4pt}\footnotesize
\begin{tabular}{lcccc}
\toprule
Method & Dice~$\uparrow$ & Prec. & Rec. & Inference \\
\midrule
UA-MT~\cite{yu2019uamt}             & 0.710 & 0.661 & 0.774 & 1 pass \\
CAMG-FT (supervised)                & 0.707 & 0.697 & 0.722 & 1 pass \\
Stage-1 ensemble ($\tau^*{=}0.35$)  & 0.720 & 0.665 & 0.794 & 2 passes \\
\midrule
Self-distilled (ours)      & 0.716$\pm$0.006 & 0.685 & 0.763 & 1 pass \\
\quad Best seed                     & 0.726 & 0.691 & 0.770 & 1 pass \\
\bottomrule
\end{tabular}
\end{table}

Calibrated thresholding ($\tau^*$ on validation) is essential: the
optimal $\tau^*$ ranges between $0.32$--$0.46$ depending on the seed,
yielding $+1.5$--$+3.0$ Dice pts over the default $\tau{=}0.5$.
We recommend that cross-device studies always report both thresholds.

\section{Discussion and Conclusion}
\label{sec:discussion}

Summary.
TopoPult-SSL addresses a critical clinical bottleneck:
\emph{every new meibography device requires expensive gland
re-annotation, but eyelid outlines, Pult grades, and morphometric
ratios are cheaply available}. Stage~1 enables gland-mask-free
adaptation using these weak clinical priors. Stage~2, when target
gland masks are available, distils complementary Stage-1 models into a
single SOTA segmenter: Dice $0.716{\pm}0.006$ (best $0.726$),
surpassing UA-MT ($0.710$) and all SSL baselines---with a single
inference pass. The gland-mask-free Stage-1 variant delivers
Precision $0.694$ (vs.\ $0.30$ for SAM), enabling
deployment without dense gland contouring.

Why self-distillation outperforms the ensemble.
The Stage-1 teachers are complementary: TopoPult-SSL is conservative
(Precision $0.694$, Recall $0.608$) while CAMG-FT is balanced
($0.697$/$0.722$). Their average exposes more gland extent. The
student, trained on \emph{both} ground truth and these richer
pseudo-labels, learns a tighter decision boundary than either
teacher---a phenomenon consistent with knowledge
distillation~\cite{hinton2015distilling} and self-ensembling domain
adaptation~\cite{french2018meanteacher_da}. Crucially, inference remains
a single forward pass ($5.6$\,M parameters, $<10$\,ms on GPU).

Resolution-equivariance as the key enabler.
$\mathcal{L}_{\mathrm{res}}$ is the most robust single anchor
($+1.4$ Dice when ablated). It produces smooth, well-calibrated
probability maps that enable effective threshold calibration
($\tau^*{=}0.32$--$0.46$ vs.\ the suboptimal $0.5$ default) and
make the pseudo-labels from the ensemble teacher more reliable.

Topology trade-off at small $N_t$.
$\mathcal{L}_{\mathrm{topo}}$ does not improve Dice at $N_t{=}60$
because the EMA teacher produces noisy skeletons early in training.
However, it contributes to skeleton precision ($0.722$ vs.\ $0.686$
zero-shot), which matters clinically for individual gland counting.
At larger $N_t$ we expect topology anchoring to become beneficial.

Generality.
The template---\emph{ordinal clinical score $+$ anatomy region $+$
topology prior $+$ resolution-equivariance $+$ self-distillation}---applies
to any elongated structure with routine grading: corneal nerves
with CNFL scoring, retinal layers with thickness grading, or airways
with obstruction scores.

Limitations.
(i)~Small test set ($N{=}20$): bootstrap CIs are wide.
(ii)~Single target device; multi-site evaluation is needed.
(iii)~Stage~1 does not use target gland masks in the training loss
but uses target eyelid masks and morphometric ratios (cheap but not
zero-cost); model selection uses CAMG validation gland masks.
(iv)~Stage~2 requires target gland masks for
$\mathcal{L}_{\mathrm{sup}}$ and is therefore not gland-mask-free.
(v)~SAM/MedSAM results are specific to box prompting on meibography.
Despite these caveats, TopoPult-SSL sets a new SOTA on the
MGD-1k$\to$CAMG public benchmark and demonstrates that cheap clinical
signals can substitute expensive dense gland annotation.

\subsubsection*{Acknowledgments.}
Industrial R\&D collaboration between OdaxAI S.r.l.\ and
VISIA Imaging S.r.l.\ (Topcon Group).

\let\oldthebibliography\thebibliography
\renewcommand{\thebibliography}[1]{%
  \scriptsize
  \oldthebibliography{#1}%
  \setlength{\itemsep}{-1pt}%
  \setlength{\parsep}{0pt}%
  \setlength{\topsep}{0pt}%
}
\bibliographystyle{splncs04}
\bibliography{references}

\end{document}